\documentclass[sigconf]{acmart}
\usepackage{bm}
\usepackage{multirow}
\usepackage{makecell}
\usepackage{url}
\usepackage{subfigure}
\usepackage{appendix}
\usepackage{stfloats}

\AtBeginDocument{%
  }

\acmDOI{}

\acmConference[DLG-KDD 2022]{Make sure to enter the correct
  conference title from your rights confirmation emai}{August 14,
  2022}{}
\acmPrice{}
\acmISBN{}

\begin{document}
\fancyhead{}
%%
%% The "title" command has an optional parameter,
%% allowing the author to define a "short title" to be used in page headers.
\title{Graph Property Prediction on Open Graph Benchmark: \\A Winning Solution by Graph Neural Architecture Search}

%%
%% The "author" command and its associated commands are used to define
%% the authors and their affiliations.
%% Of note is the shared affiliation of the first two authors, and the
%% "authornote" and "authornotemark" commands
%% used to denote shared contribution to the research.

\author{Xu Wang}
\affiliation{%
  \institution{4Paradigm}
  % \streetaddress{Address}
  \city{Beijing}
  \country{China}}
\email{wangxu01@4paradigm.com}

\author{Huan Zhao}
\affiliation{%
  \institution{4Paradigm}
  % \streetaddress{Address}
  \city{Beijing}
  \country{China}}
\email{zhaohuan@4paradigm.com}

\author{Lanning Wei}
\affiliation{%
  \institution{4Paradigm}
  % \streetaddress{Address}
  \city{Beijing}
  \country{China}}
\email{weilanning18z@ict.ac.cn}

\author{Quanming Yao}
\affiliation{%
  \institution{Department of Electronic Engineering, Tsinghua University}
  % \streetaddress{Address}
  \city{Beijing}
  \country{China}}
\email{qyaoaa@tsinghua.edu.cn}

% \author{Lars Th{\o}rv{\"a}ld}
% \affiliation{%
%   \institution{The Th{\o}rv{\"a}ld Group}
%   % \streetaddress{1 Th{\o}rv{\"a}ld Circle}
%   \city{Hekla}
%   \country{Iceland}}
% \email{larst@affiliation.org}

%%
%% By default, the full list of authors will be used in the page
%% headers. Often, this list is too long, and will overlap
%% other information printed in the page headers. This command allows
%% the author to define a more concise list
%% of authors' names for this purpose.
% \renewcommand{\shortauthors}{Trovato et al.}

%%
%% The abstract is a short summary of the work to be presented in the
%% article.
\begin{abstract}
% \xu{Replace the relevant description of automl with NAS and modify the mentioned statements. }

  % OGB (Open Graph Benchmark) emerge as the times require. 
  % It integrates datasets of various sizes from different fields and provides different tasks on graphs.
  % As a cutting-edge branch of machine learning, NAS (Neural Architecture Search) has proved its generalization ability in many fields.
  Aiming at two molecular graph datasets and one protein association subgraph dataset in OGB graph classification task, we design a graph neural network framework for graph classification task by introducing PAS(Pooling Architecture Search).  
  % \wei{design a graph neural network framework for graph classification task by xxx?}
  At the same time, we improve it based on the GNN topology design method $\text{F}^{2}\text{GNN}$ to further design the feature selection and fusion strategies, so as to further improve the performance of the model in the graph property prediction task while overcoming the over smoothing problem of deep GNN training.
  Finally, a performance breakthrough is achieved on these three datasets, which is significantly better than other methods with fixed aggregate function. 
  It is proved that the NAS method has high generalization ability for multiple tasks and the advantage of our method in processing graph property prediction tasks.
\end{abstract}

%% A "teaser" image appears between the author and affiliation
%% information and the body of the document, and typically spans the
%% page.

%%
%% This command processes the author and affiliation and title
%% information and builds the first part of the formatted document.
\maketitle

\section{Introduction}
\begin{figure}[t]
  \centering
  \includegraphics[width=\linewidth]{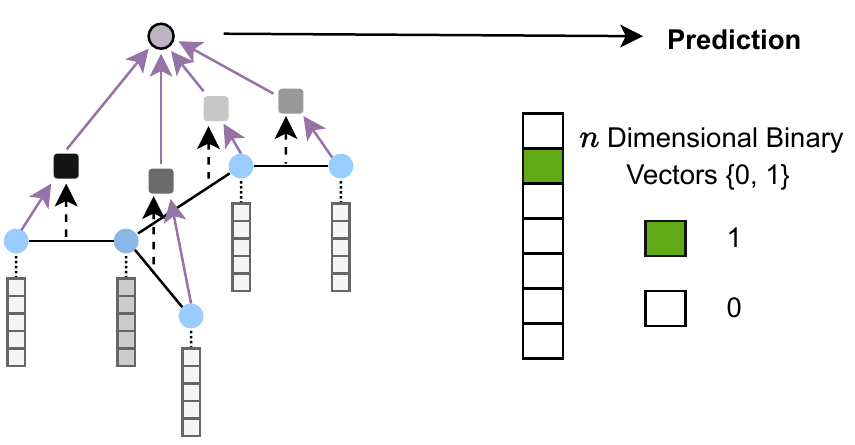}
  \caption{Graph property prediction task of molecular samples. Blue circles represent atomic nodes, gray rectangles represent node features, and purple circles represent graph-level embedding extracted by graph neural network.}
  \label{fig:moleGraph}
\end{figure}

Graph structure has been widely used in various abstract interactive complex systems, such as social networks \cite{easley2010networks}, knowledge graphs \cite{nickel2015review}, molecular diagrams \cite{wu2018moleculenet} and biological networks \cite{barabasi2004network, wang2022graph}.
Recently, a series of progress has been made in the application of graph related machine learning methods in various fields \cite{xu2018powerful}.
In order to further promote the research of machine learning on graphs, Weihua Hu team of Stanford University put forward the open graph benchmark dataset in NeurIPS 2020 \cite{hu2020open}, which is committed to promoting the research of robust and reproducible graph machine learning.
OGB dataset involves many fields, and the tasks are divided into node-level, edge-level and graph-level prediction tasks.
Among them, the graph-level prediction task includes four datasets from three different application fields, including two molecular graph datasets of \textsl{ogbg-molhiv} and \textsl{ogbg-molpcba}, protein-protein association subgraph dataset \textsl{ogbg-ppa}, and \textsl{ogbg-code}, a set of source code of Python abstract syntax tree.
The graph-level prediction of molecular properties is shown in Figure~\ref{fig:moleGraph}. The graph neural network is used to extract information from the atomic and chemical bond features stored on nodes and edges. The representation of the whole graph is obtained according to the updated representation of each node, so as to convert the prediction of molecular properties into graph-level classification.

Based on the first three datasets of OGB graph classification task, we design a task adaptive graph neural network architecture combined with neural architecture search (NAS) \cite{zoph2016neural,zhao2021search,wei2021pooling}  on graph, so as to achieve further performance breakthrough on each corresponding task.
Specifically, we introduce the pooling structure search scheme PAS (pooling architecture search) \cite{wei2021pooling}, which designs the aggregate function search space, and applies the differentiable search algorithm to the specific aggregate function and pooled structure search.
At the same time, we also improve PAS based on the graph neural network topology design method $\text{F}^{2}\text{GNN}$ (Feature Fusion GNN) \cite{wei2022designing}, which designs the topology of graph neural network from the perspective of differentiable search method and feature fusion, so as to overcome the problem of over smoothing (reduction of feature discrimination of adjacent nodes) in the training of deep graph neural network, and finally obtain $1$st place on the first three datasets, i.e., \textsl{ogbg-molhiv}, \textsl{ogbg-molpcba}, and \textsl{ogbg-ppa}, in the graph-level prediction task.\footnote{Note in the second dataset, i.e., \textsl{molpcba}, there are two settings. We obtain the $1$st place among those do not use external dataset.} Code is released in: \url{https://github.com/AutoML-Research/PAS-OGB}.

\section{Background}
\subsection{OGB Graph Property Prediction}
% \textsl{ogbg-molhiv} and \textsl{ogbg-molpcba} come from MoleculeNet \cite{wu2018moleculenet} and are one of the largest datasets, and they contain more than 40,000 and 400,000 molecular samples respectively.
% This task requires to construct the graph structure with atoms as nodes and chemical bonds as edges, and predict the characteristics of each molecular sample on this basis, as shown in Figure~\ref{fig:moleGraph}. The former focuses on the ability of molecules to inhibit HIV virus replication, and the latter focuses on the effectiveness of different compounds against more than 100 disease targets, which is of great significance for the screening and development of effective drugs for many diseases.

\textsl{ogbg-molhiv} and \textsl{ogbg-molpcba}, provided by MoleculeNet \cite{wu2018moleculenet}, are the molecule property prediction datasets, where nodes are the atoms and the edges represents the chemical bonds in each molecule graph.
The former focuses on the ability of molecules to inhibit HIV virus replication, and the latter focuses on the effectiveness of different compounds against more than 100 disease targets. These two datasets contain more than 40,000 and 400,000 molecular samples, respectively, which are the largest molecule property datasets in \cite{wu2018moleculenet}. Therefore, it is greatly significant to utilize these two datasets to screen and develop effective drugs for many diseases.

\textsl{ogbg-ppa} dataset is a set of undirected protein association subgraphs extracted from the protein association networks of 1,581 different species, covering 37 broad biological taxa (including mammals, bacteroideae, archaea, etc.), including more than 150,000 samples. The task requires to give a protein association subgraph and predict the source of biological taxa of the association subgraph. Successfully solving this problem is of great significance to understand the evolution of cross species protein complexes, the recombination of protein interactions over time, and the discovery of functional associations between genes. It can also have an in-depth understanding of key bioinformatics tasks, such as biological network alignment. Further details of the three datasets are shown in the appendix.

At present, a variety of GNNs have been applied to graph classification task, but there is not a general method on the three datasets that can stably achieve SOTA performance.
In addition, graph-level representation learning may require long-range neighbors~\cite{wu2021representing,xu2018representation}, and the top schemes for two molecular datasets on the leaderboard mostly adopt deeper GNNs. 
However, the deep GNN model often leads to the over smoothing problem of reducing the feature distinguishability of neighbor nodes and significantly reducing the final prediction effect.

\subsection{Graph Neural Architecture Search}
%Neural architecture search (NAS) is an important direction in the field of AutoML, and it aims to find the neural network structure of SOTA on a certain task or a certain kind of task in the predefined search space. At present, the existing representative methods include \cite{li2020sgas, liu2018darts, xie2018snas, zoph2016neural}.
%At the same time, with the popularization of GNN, there are also attempts to automatically design GNN network structure through NAS. Most of these methods focus on the design of aggregation function. 

Neural architecture search (NAS) methods were proposed to automatically find SOTA CNN architectures in a pre-defined search space and representative methods are~\cite{li2020sgas, liu2018darts, xie2018snas, zoph2016neural}. 
At the same time, most of the existing GNN designs follow the message passing framework to design the aggregatrion functon of neighbor information and the update function of node representation \cite{gilmer2017neural}.
Very recently, researchers tried to automatically design GNN architectures by NAS. 
The majority of these methods focus on designing the aggregation layers in GNNs.  
For example, the representative method GraphNAS \cite{gao2020graph} defines a search space that includes an attention function, the number of attention heads, and embedded size, etc. Based on the JKNet \cite{xu2018representation}, SNAE \cite{zhao2020simplifying} and SNAG \cite{zhao2021search} 
learn to select and fuse the characteristics in each layer.
%select and fuse the characteristics of the final output node and each middle layer.
%
Following the paradigm of NAS, one search algorithm should be employed to select architetcures from the designed search space.
%
%In addition to designing search space, the design of search strategy is also the key aspect in NAS, such as 
Some work that directly introduces reinforcement learning or evolutionary algorithm into search strategy \cite{gao2019graphnas, zhao2020simplifying, chen2021graphpas, guo2020single}.
In order to further improve the search efficiency, some methods relax the discrete search space into a continuous search space, and then they can be optimized with the gradient descent, which is efficient in orders of magnitude than the aforementioned ones.
%defines the neural network architecture search as a bi-level optimization problem \cite{cai2021rethinking, li2021one, liu2018darts}. 

PAS designs the pooling architectures, including \textbf{Aggregation}, \textbf{Pooling}, \textbf{Readout} and \textbf{Merge} operations, which are essential for obtaining graph-level representations. Then it developed an effective differentiable search algorithm by a continuous relaxation of the search space. And in this way, data-specific architectures are obtained. $\text{F}^{2}\text{GNN}$ provide a feature fusion perspective in designing GNN topology and propose a novel framework to unify the existing topology designs with feature selection and fusion strategies. 
In this paper, we transform the search space of PAS scheme, modifies the search space of aggregation function while deleting the pooling operation search, and introduces the neural architecture search method of $\text{F}^{2}\text{GNN}$ to design network topology, so as to further improve the effect of adapting to downstream tasks after feature selection and fusion.

\section{Method}

In this section, we elaborate on the proposed novel framework PAS, which is based on NAS to search for adapative architectures for graph classification, consisting of the novel search space and the efficient search algorithm. At the same time, we will also elaborate on the improvement scheme of PAS based on $\text{F}^{2}\text{GNN}$ method.

\noindent\textbf{Notations.}
We represent a graph as $G =(\textbf{A}, \textbf{H})$, where $\textbf{A} \in \mathbb{R}^{N \times N}$ is the adjacency matrix of this graph and $\textbf{H} \in \mathbb{R}^{N \times d}$ is the node features, $N$ is the node number and $d$ is the feature dimension. For simplicity, all the intermediate layers have the same feature dimension $d$. The input graph is represented as $G =(\textbf{A}, \textbf{H}^0)$ in this paper.

\subsection{The Design of the Search Space}
\begin{figure}[h]
  \centering
  \includegraphics[width=\linewidth]{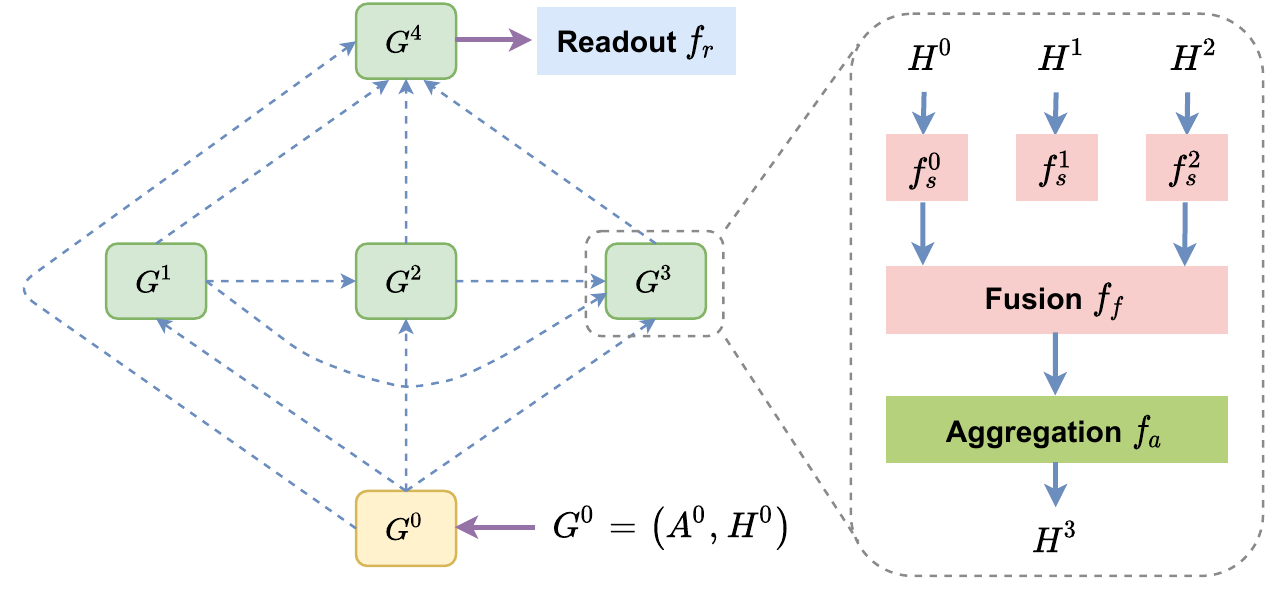}
  \caption{The framework of our method. For the $i-$th SFA block, we have $i$ selection operations $f_{s}$ and one fusion operation $f_{f}$ to select and integrate the output of the previous block. Then one aggregation operation $f_{a}$ is followed to aggregated messages from the neighborhood. The last SFA block will be followed by a readout operation $f_{r}$ to get graph representation.}
  \label{fig:pas+f2gnn}
\end{figure}

As shown in Figure~\ref{fig:pas+f2gnn}, the proposed framework contains $N$ SFA (Selection, Fusion and Aggregation) blocks ($N=4$ for example) to extract the local information, and then adds one \textbf{Readout} module to generate the graph representations for downstream tasks.
For the $i-$th SFA block, \textbf{Selection} module determines whether to use the output characteristics of the first $i$ block. \textbf{Fusion} module determines the method of intergrating these selected features. The \textbf{Aggregation} operation $f_{a}$ update the representation of each node without changing the structure of the graph itself. Therefore, the high-level features can be generated by $\textbf{H}^i = f_a(f_f(\{f_s^0(\textbf{H}^0), \cdots, f_s^{i-1}(\textbf{H}^{i-1})\}))$.
Here, $f_s$ is the selection operation and $f_f$ is the fusion operation.
At the end of the SFA blocks, we add one readout operation to generate the graph representations based on the aggregation results.

\begin{table}
  \caption{The operations used in our search space.}
  \label{tab:search-space}
  \begin{tabular}{c|c}
    \toprule
    Module & Operations\\
    \hline
    Selection $O_{s}$ & $\textbf{ZERO}$, $\textbf{IDENTITY}$\\ \hline
    Fusion $O_{f}$ & \textbf{SUM}, \textbf{MEAN}, \textbf{MAX}, \textbf{CONCAT}, \textbf{LSTM}\\  \hline
    \multirow{2}{*}{Aggregation$O_{a}$} & \textbf{GCN}\cite{kipf2016semi}, \textbf{GAT}\cite{velivckovic2017graph}, \textbf{GIN}\cite{xu2018powerful}, \textbf{GEN}\cite{li2020deepergcn}, \\ & \textbf{MF}\cite{duvenaud2015convolutional}, \textbf{ExpC}\cite{DBLP:journals/corr/abs-2012-07219}\\ \hline
    \multirow{2}{*}{Readout $O_{r}$}  & \textbf{GLOBAL\_MEAN}, \textbf{GLOBAL\_MAX}, \\ & \textbf{GLOBAL\_SUM} \\
  \bottomrule
\end{tabular}
\end{table}

The candidate operations we used in the search space are provided in Table~\ref{tab:search-space}. Following $\text{F}^{2}\text{GNN}$, two operations IDENTITY and ZERO are provided in the Selection module, which represents the ``selected`` and ``not selected`` stages for the input feature. Five fusion operations are selected to fuse these features with the summation, average, maximum, concatenation and LSTM cell, which are denoted as SUM, MEAN, MAX, CONCAT and LSTM, respectively.
For Aggregation module, apart from the widely used aggregation operations GCN~\cite{kipf2016semi}, GAT~\cite{velivckovic2017graph} and GIN~\cite{xu2018powerful}, we further provide three operations GEN\cite{li2020deepergcn},  MF\cite{duvenaud2015convolutional} and ExpC\cite{DBLP:journals/corr/abs-2012-07219} which achieved great progress in the OGB leaderboard.
Following PAS, we take the maximum, mean and summation of all node embeddings as the graph-level representations, and these three operations are denoted as GLOBAL\_MEAN, GLOBAL\_MAX and GLOBAL\_SUM, respectively.
% \xu{The explanation of deleting the pooling operation written by lanning is adopted.}

\noindent\textbf{Discussions.}
Considering that the OGB datasets \textsl{ogbg-molhiv} and \textsl{ogbg-molpcba} are the molecule datasets where the graph structures are largely correlated with the graph property, pooling operations aiming to generate the coarse graphs may lead to information loss to some extent for these two datasets~\cite{grattarola2021understanding}. Besides, the top schemes in the leaderboard also ignore the pooling operations. Therefore, compared with PAS, we delete the \textbf{Pooling} module and turn to designing deep GNNs to capture the long-range dependencies in OGB datasets.
In this paper, we focus on extracting the long-range dependencies for the graph classification task based on the deep GNNs rather than the pooling operations.
Considering the over smoothing problem of deep GNN network training, we introduce the  $\text{F}^{2}\text{GNN}$ which can alleviate this problem by utilizing the features in different ranges adaptively. 
This method is designed for the node classification task, and we extend it into the graph classification task which is a natural application of deep GNNs.

\subsection{The Design of the Search Algorithm}
In order to make the search algorithm continuously differentiable, we design the following relaxation function to relax the discrete search space into a continuous space in the form of weighted summation:
\begin{equation}
  \bar{o}(x)=\sum_{i=1}^{\left |  O \right | }c_{i}o_{i}(x),  
\end{equation}
where $c_{i}\in(0,1)$ is the weight of the $i-$th operation $o(\cdot)$ in set $O$. Among them, the selection module contains two opposite operations: \textbf{ZERO} stands for not selecting the feature, while \textbf{IDENTITY} stands for selecting the input feature.
Therefore, the mixed selection result of this module can be expressed as:
\begin{equation}
  \bar{o}^{ij}(x_{i})= {\textstyle \sum_{k=1}^{\left | O_{s} \right | }} c_{k}^{ij}o_{k}^{ij}(x_{i})=c_{1}^{ij}0+c_{2}^{ij}x_{i}=c_{2}^{ij}x_{i},
\end{equation}
since the weight of \textbf{ZERO} operation will be fixed multiplied by 0, there will be an error in the final result, and the error will gradually expand with the advancement of the calculation process.
Therefore, we introduce a temperature coefficient $\lambda$ when calculating the operation weight, and make the weight coefficient of each operation expressed as: 
\begin{equation}
  c_{k} = \frac{exp(\alpha_{k} /\lambda )}{ {\textstyle \sum_{i=1}^{\left | O \right | }}exp(\alpha_{i}/\lambda) },
\end{equation}
where $\alpha_{i}$ is the supernet parameter corresponding to operation $o_{i}(\cdot)$. When a smaller temperature coefficient is set, the operation weight tends to be close to a one-hot vector, so as to reduce the influence between each other.
\section{Experiment}
All models are implemented with Pytorch (version 1.8.0) on a GPU RTX3090. In addition, in order to facilitate the implementation of various GNN variants, we use the popular GNN library: PYG \cite{fey2019fast} (Pytorch Geometric) (version 2.0.1).
The experimental hyper-parameter settings on the three datasets are shown in appendix.

Most of the baselines we selected are within 5 in the OGB leaderboards (the experiments on the \textsl{ogbg-molpcba} do not include the introduction of extend datasets for pre-training).
For the two molecular property prediction tasks, we use binary cross-entropy as the loss function, while for the multi-class classification task for \textsl{ogbg-ppa}, we use cross-entropy as the loss function.

\subsection{ogbg-molhiv}
For \textsl{ogbg-molhiv}, we automatically designed a 14 layer GNN model according to the average diameter of the graph in its dataset, and introduced two molecular fingerprints: MorganFingerprint and MACCSFingerprint \cite{Molecule_fingerprint}.
We jointly train PAS and molecular fingerprint model based on random forest, and use softmax and a learnable parameter to aggregate the two results. Considering the advantages of DeepAUC \cite{yuan2020large} in binary classification task with unbalanced samples, we introduce it into the calculation of loss function. The searched architecture is shown in Figure~\ref{fig:arch-hiv}, it mainly includes GEN, updates the node representation in deeper GNN layer, and uses MF to combine the node degree in the last three layers, which reflects the importance of atomic degree for this task. The final experimental results are shown in Table~\ref{tab:hiv-res}.
It can be seen that compared with other fixed aggregation function schemes, PAS can automatically search the aggregation function of each layer in GNN, and has achieved SOTA on this dataset.

\subsection{ogbg-molpcba}

The sample size of \textsl{ogbg-molpcba} is about ten times larger than that of \textsl{ogbg-molhiv}. Preliminary experiments show that the attention mechanism has a good performance in this task,  we updated the search space in the aggregation function to several variants of GAT \cite{zhao2021search}, and designed an 18 layer GNN model combined with the average diameter of the graph in the dataset.
The searched architecture after introducing $\text{F}^{2}\text{GNN}$ is shown in Figure~\ref{fig:arch-pcba}, compared with the connect operation of dense \cite{huang2018condensenet}, our method is more flexible in obtaining higher-level features, and interms of effect, the final performance is shown in the Table~\ref{tab:pcba-res}. Compared with PAS, the performance on \textsl{ogbg-molpcba} has been significantly improved after the introduction of architecture search, and the effect of SOTA without introducing additional datasets confirm the effectiveness of the scheme in the prediction of molecular graph properties.

\begin{table}
  \caption{Performance on ogbg-molhiv dataset. ROC-AUC is used as the metric of this binary-class classification task.}
  \label{tab:hiv-res}
  \begin{tabular}{ccc}
    \toprule
    Method & Test AUC & Validation AUC\\ \hline
    Graphormer + FPs \cite{ying2021transformers} & $0.8225 \pm 0.0001$ & $0.8396 \pm 0.0001$\\
    CIN \cite{bodnar2021weisfeiler} & $0.8094 \pm 0.0057$ & $0.8277 \pm 0.0099$\\
    GMAN + FPs \cite{bag_of_trick}  & $0.8244 \pm 0.0033$ & $0.8329 \pm 0.0039$\\ 
    DeepAUC \cite{yuan2020large} & $0.8352 \pm 0.0054$ & $0.8238 \pm 0.0061$ \\ \hline
    \textbf{PAS+FPs} & $\textbf{0.8420} \pm \textbf{0.0015}$ & $\textbf{0.8238} \pm \textbf{0.0028}$ \\
    \bottomrule
\end{tabular}
\end{table}
\begin{table}
  \caption{Performance on ogbg-molpcba dataset. Average Precision (AP) is used as the metric of this multiple bianry classification task.}
  \label{tab:pcba-res}
  \begin{tabular}{ccc}
    \toprule
    Method & Test AP & Validation AP\\ \hline
    GIN \cite{xu2018powerful} & $0.2834 \pm 0.0038$ & $0.2912 \pm 0.0026$\\
    CRaWl  \cite{toenshoff2021graph} & $0.2986 \pm 0.0025$ & $0.3075 \pm 0.0020$\\
    GINE + bot \cite{bag_of_trick}  & $0.2994 \pm 0.0019$ & $0.3094 \pm 0.0023$\\ 
    Nested GIN \cite{zhang2021nested} & $0.3007 \pm 0.0037$ & $0.3059\pm 0.0056$ \\ \hline
    \textbf{PAS} & $\textbf{0.3012} \pm \textbf{0.0039}$ & $\textbf{0.3151} \pm \textbf{0.0047}$ \\
    \textbf{PAS+$\text{F}^{2}\text{GNN}$} & $\textbf{0.3147} \pm \textbf{0.0015}$ & $\textbf{0.3258} \pm \textbf{0.0017}$ \\
    \bottomrule
\end{tabular}
\end{table}
\begin{table}
  \caption{Performance on ogbg-ppa dataset.}
  \label{tab:ppa-res}
  \begin{tabular}{ccc}
    \toprule
    Method & Test Accuracy & Validation Accuracy\\ \hline
    DeeperGCN \cite{li2020deepergcn} & $0.7752 \pm 0.0069$ & $0.7484 \pm 0.0052$\\
    PAS & $0.7828 \pm 0.0024$ & $0.7523 \pm 0.0028$\\ 
    ExpC  \cite{DBLP:journals/corr/abs-2012-07219} & $0.7976 \pm 0.0072$ & $0.7518 \pm 0.0080$\\
    ExpC* \cite{bag_of_trick} & $0.8140 \pm 0.0028$ & $0.7811\pm 0.0012$ \\ \hline
    \textbf{PAS+$\text{F}^{2}\text{GNN}$} & $\textbf{0.8201} \pm \textbf{0.0019}$ & $\textbf{0.7720} \pm \textbf{0.0023}$ \\
    \bottomrule
\end{tabular}
\end{table}

\subsection{ogbg-ppa}
The \textsl{ogbg-ppa} dataset has no initial node characteristics, and its graph specification is about 10 times higher than the first two datasets, and the training cycle is significantly increased.
Considering the good performance of ExpC \cite{DBLP:journals/corr/abs-2012-07219} in the graph property prediction task without initial node characteristics, we fixed it in the aggregation module, and design GNN topology based on SFA block.

\begin{figure}[t]
  \centering
  \subfigure[Architecture on \textsl{ogbg-molhiv}]{
    \includegraphics[width=0.5\linewidth]{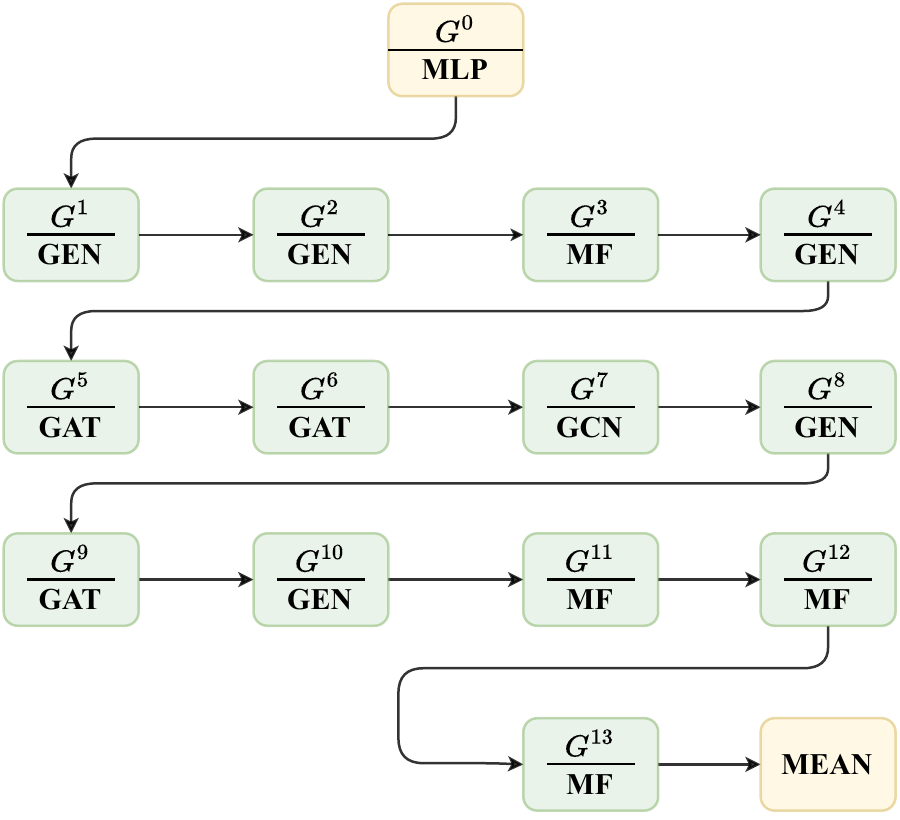}
    \label{fig:arch-hiv}
  }
  \subfigure[Architecture on \textsl{ogbg-ppa}]{
    \includegraphics[width=0.35\linewidth]{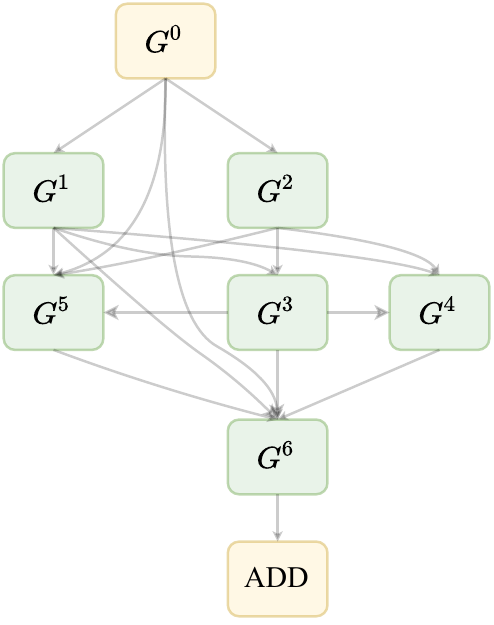}
    \label{fig:arch-ppa}
  }
  \\
  \subfigure[Architecture on \textsl{ogbg-molpcba}]{
    \includegraphics[width=0.55\linewidth]{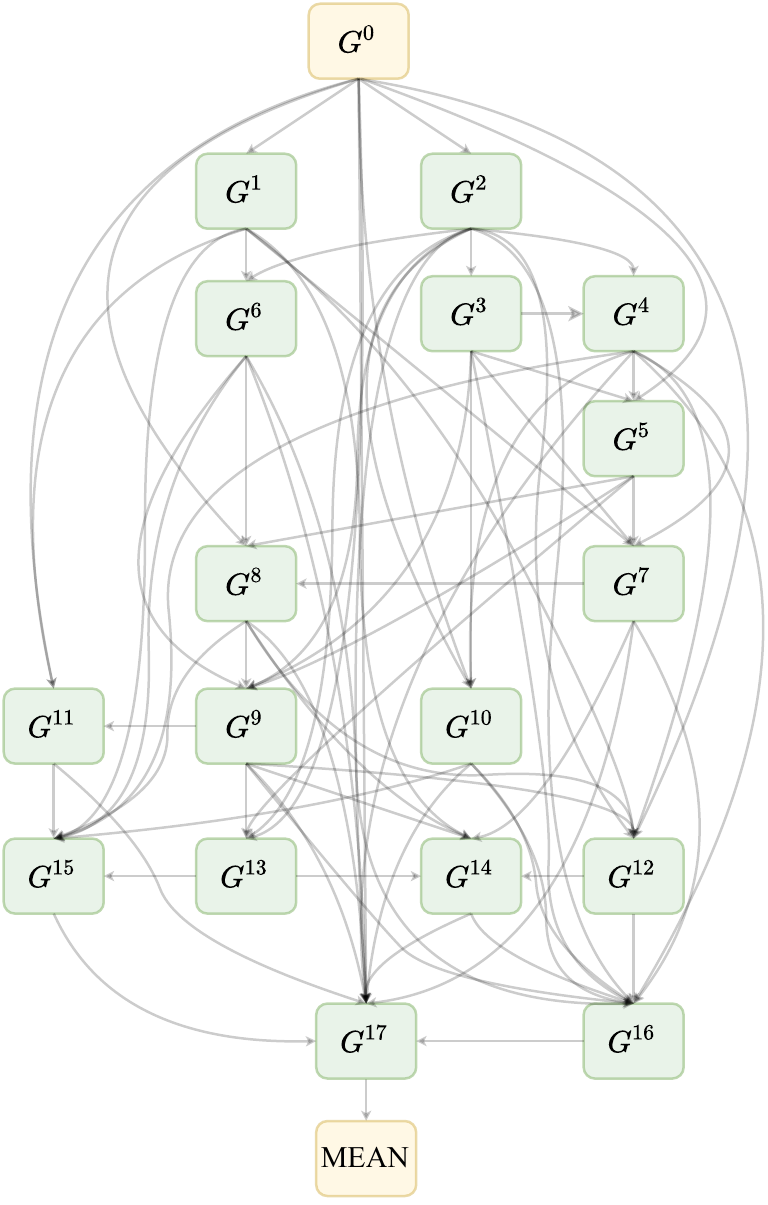}
    \label{fig:arch-pcba}
  }

  \caption{The searched results on ogbg-molhiv, ogbg-molpcba and ogbg-ppa datasets. The last yellow blocks indicate readout operations. (a) For ogbg-molhiv, the result of aggregate functions searched for PAS, not involving topology search. (b) For ogbg-ppa, fixed the aggregate function to ExpC and search the topology. (c) For ogbg-molpcba, fixed the aggregate function to the variety of GAT and search the topology.}
  
\end{figure}

The searched architecture after introducing $\text{F}^{2}\text{GNN}$ is shown in Figure~\ref{fig:arch-ppa},
and the final performance on \textsl{ogbg-ppa} is shown in the Table~\ref{tab:ppa-res}. After introducing the topology search based on f2gnn, this scheme is significantly higher than the original schemes of expc and PAS, which verifies the importance of GNN topology and the advantages of topology adaptive design of this scheme.

\section{Conclusion}
For the three datasets of OGB graph classification task, we build our method based on two existing neural architecture search methods, i.e., pooling architecture search PAS, and GNN topology search $\text{F}^{2}\text{GNN}$. The results show that our approach has made a performance breakthrough in three tasks, which demonstrates the advantages of NAS methods for GNN.
For future work, we will further explore the topology search problem of deeper GNN framework, and further apply the method to real-world, especially bioinformatic, applications.

%%
%% The next two lines define the bibliography style to be used, and
%% the bibliography file.
\bibliographystyle{ACM-Reference-Format}
\bibliography{sample-base}

\clearpage
\begin{appendices}
  \section{Task detail and experiment settings}
  The specific details of OGB graph property prediction datasets are shown in the Table~\ref{tab:ogb}.
  \textsl{ogbg-molhiv} is a single binary-class classification task, and ROC-AUC(Receiver Operating Characteristic-Area Under the Curve) is used as the evaluation metric.
  \textsl{ogbg-molpcba} is a multiple binary-class classification task, with AP(Average Precision) as the metric.
  The last dataset \textsl{ogbg-ppa} is a multy-class classification task, and the diameter rises sharply. All graph samples have no node features, only edge features, and accuracy is used as the evaluation metric.

  The hyperparametric search space of our method on three datasets is shown in the Table~\ref{parameter}. It should be noted that gamma is a hyper-parameter of the introduced DeepAUC \cite{yuan2020large}. This method is only applicable to binary classification tasks, so gamma only appears in ogbg-molhiv dataset.

  \begin{table*}
    \caption{Introduction of OGB Graph Property Prediction Datasets.}
    \label{tab:ogb}
    \begin{tabular}{ccccccc}
      \toprule
      Dataset&\# Graphs&\# Nodes per graph&\# Edges per graph&\# Tasks&Task Type&Metric\\
      \midrule
      \textsl{ogbg-molhiv} & 41,127& 25.5&27.5&1&Binary classification&ROC-AUC\\
      \textsl{ogbg-molpcba} & 437,929& 26.0&28.1&128&Binary classification&AP\\
      \textsl{ogbg-ppa}& 158,100 & 243.4&2,266.1&1&Multi-class classification&Accuracy\\
    \bottomrule
  \end{tabular}
  \end{table*}

\begin{table}[h]
  \caption{Hyper-parameter search space.}
  \label{parameter}
  \centering
  \begin{tabular}{cccc}
  \toprule
   & ogbg-molhiv & ogbg-molpcba & ogbg-ppa \\
  \midrule
    \makecell[c]{learning \\ rate}&\makecell[c]{[5e-3, 1e-2, \\ 3e-2, 5e-2, \\ 1e-1]} &\makecell[c]{[5e-4, 1e-3, \\ 3e-3, 5e-3, \\ 1e-2]} &\makecell[c]{[5e-3, 1e-2, \\ 3e-2, 5e-2, \\ 1e-1]} \\
    \midrule
    batch size & \makecell[c]{[128, 256, \\ 512]} & \makecell[c]{[256, 512, \\ 1024]}& \makecell[c]{[128, 256, \\ 512]} \\
    \midrule
    hidden size & [256, 512] & [512, 1024]& [256, 512] \\
    \midrule
    dropout & [0.1, 0.2, 0.3] & [0.1, 0.2, 0.3]& [0.1, 0.2, 0.3] \\
    \midrule
    gamma & [500, 700, 1000] & [-] & [-] \\
    \midrule
    virtual node & [True, False] & [True, False] & [True, False] \\
    
    % batch_size & [128, 256, 512] & [256, 512, 1024] & [128, 256, 512] \\
    \bottomrule
  \end{tabular}
\end{table}

\end{appendices}
\end{document}